\documentclass[a4paper,conference]{IEEEtran}

\usepackage[utf8]{inputenc}
\usepackage[T1]{fontenc}
\usepackage{hyperref}
\usepackage{url}
\usepackage{booktabs}
\usepackage{amsfonts}
\usepackage{nicefrac}
\usepackage{microtype}
\usepackage{xcolor}
\usepackage{graphicx}
\usepackage{subcaption}
\usepackage{amsmath,amsfonts,bm}
\usepackage{wrapfig}
\usepackage{adjustbox}
\usepackage{pdfpages}
\usepackage{multido}

\newcommand{\framework}{NetSpace}
\newcommand{\prepr}{PRep}

\newcommand{\kdloss}{\mathcal{L}_{kd}}
\newcommand{\predloss}{\mathcal{L}_{pred}}
\newcommand{\predlossmulti}{\mathcal{L}_{pred}^*}
\newcommand{\taskloss}{\mathcal{L}_{task}}
\newcommand{\classloss}{\mathcal{L}_{class}}
\newcommand{\celoss}{CE}
\newcommand{\klloss}{KL}
\newcommand{\softmax}{\operatorname{softmax}}
\newcommand{\interplossclass}{\mathcal{L}_{class}^\gamma}
\newcommand{\interplosspred}{\mathcal{L}_{kd}^\gamma}
\newcommand{\interplosspredone}{\mathcal{L}_{pred}^\gamma}
\newcommand{\interplosspredtwo}{\mathcal{L}_{task}^\gamma}
\newcommand{\interploss}{\mathcal{L}^{\gamma}}
\newcommand{\classid}{ClassId}
\newcommand{\lenet}{LeNetLike}
\newcommand{\vanillacnn}{VanillaCNN}

\begin{document}

\title{Learning the Space of Deep Models}

\author{\IEEEauthorblockN{Gianluca Berardi*, \ Luca De Luigi*,  \ Samuele Salti, \ Luigi Di Stefano\\}
\IEEEauthorblockA{Department of Computer Science and Engineering (DISI), 
University of Bologna, Italy\\
{\tt\small \{gianluca.berardi3, luca.deluigi4, samuele.salti , luigi.distefano\}@unibo.it}\\
{* \tt\footnotesize Corresponding authors with equal contribution.}}
}

\maketitle

\begin{abstract}
Embedding of large but redundant data, such as images or text, in a hierarchy of lower-dimensional spaces is one of the key features of representation learning approaches, which nowadays provide state-of-the-art solutions to problems once believed hard or impossible to solve.
In this work\footnote{Code available at \url{https://github.com/CVLAB-Unibo/netspace}.}, in a plot twist with a strong meta aftertaste, we show how trained deep models are as redundant as the data they are optimized to process, and how it is therefore possible to use deep learning models to embed deep learning models. In particular, we show that it is possible to use representation learning to learn a fixed-size, low-dimensional embedding space of trained deep models and that such space can be explored by interpolation or optimization to attain ready-to-use models.
We find that it is possible to learn an embedding space of multiple instances of the same architecture and of multiple architectures.
We address image classification and neural representation of signals, showing how our embedding space can be learnt so as to capture the notions of performance and 3D shape, respectively. In the Multi-Architecture setting we also show  how an embedding trained only on a subset of architectures can learn to generate already-trained instances of architectures it never sees instantiated at training time.
\end{abstract}

\begin{figure*}
    \includegraphics[width=0.85\textwidth]{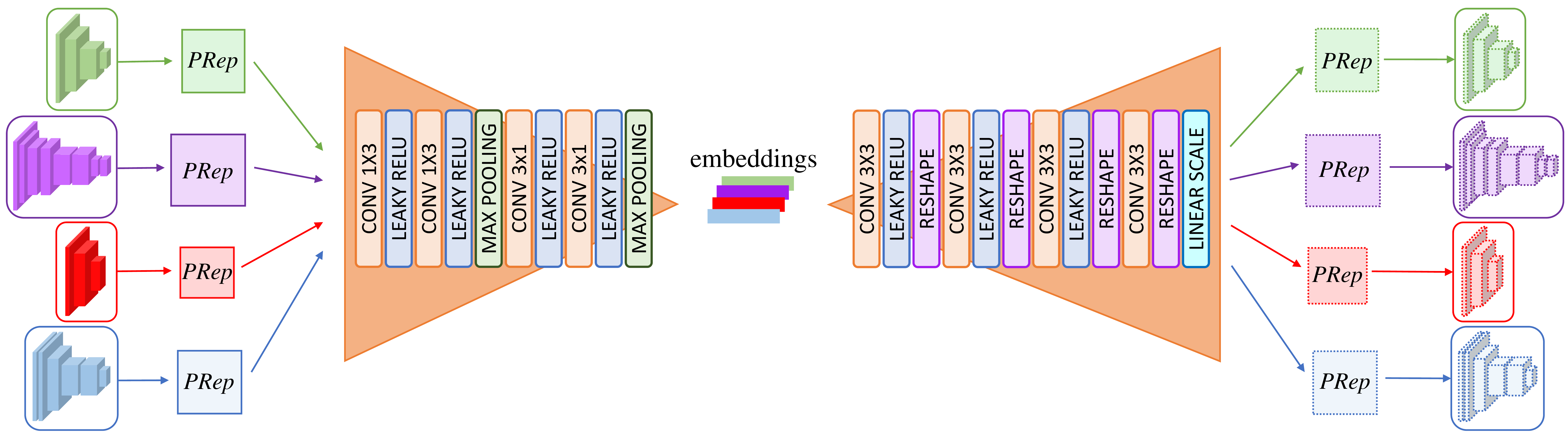}
    \centering
    \caption{Overview of the \framework{} framework.}
    \label{fig:framework}
\end{figure*}

\section{Introduction}
Representation learning has achieved remarkable results in embedding text, sound and images into low dimensional spaces, so as to map semantically close data into points close one to another into the learnt space. 
In recent years, deep learning has emerged as the most effective machinery to pursue representation learning, many scholars agreeing on representation learning laying at the very core of the deep learning paradigm.
On the other hand, the success of network compression and pruning approaches \cite{choudhary2020comprehensive} highlight the redundancy of parameters learned by a deep learning model, as in the Lottery Ticket Hypothesis \cite{frankle2018lottery}, which shows that training as few parameters as 4\% of those of the full network (i.e. the \textit{winning tickets}) can attain similar or even higher performance.

Thus, we felt puzzling and worth investigating whether the parameter values of a trained deep model might be squeezed into a semantically meaningful low-dimensional latent space.
Two questions arise: is it possible to train a deep learning model to learn to represent other, already trained, deep learning models?
And according to which trait should two already trained models lay either close or further away in the latent space? 
The Lottery Ticket Hypothesis may suggest the existence of a low-dimensional key set of information that is shared by all possible sets of parameters for a predefined architecture that achieve comparable performance on a given task. Hence, it seems reasonable to conjecture that one might pursue learning of an embedding space shaped according to similarity in performance.
Moreover, many recent works have demonstrated how small deep networks can be trained to fit accurately complex signals such as images\cite{sitzmann2020implicit}, implicit representations of 3D surfaces \cite{park2019deepsdf, mescheder2019occupancy} and even radiance fields \cite{mildenhall2020nerf}. One might then be willing to embed such models into a space amenable to capture the similarity between the underlying signals.  

In this paper we propose a first investigation along this new line of research. In particular, we show that it is  possible to deploy a basic encoder-decoder architecture to learn a low-dimensional latent space of deep models  and that such a space can be shaped so to exhibit a semantically meaningful structure. We posit that the loss to drive the learning process of our encoder-decoder architecture should entail functional similarity - rather than proximity of parameter values - between the input and output models. Accordingly, we train our architecture by knowledge distillation to drive the output model generated by the decoder to mimic the behaviour of the input model.
In our study we address two settings: learning a latent space from a training set of models with the same network architecture and different parameter values as well as based on a training set comprising models with different architectures. In both settings, we show that the learnt latent space does posses a semantic structure as it is possible to sample new trained models with predictable behaviour by simple interpolation operations. Moreover, we show that in the Multi-Architecture setting a latent space trained on a set of architectures can generate already-trained models of architectures never seen instantiated at training time. Finally, we show that in both settings it is possible to train an architecture by performing latent space optimization on the low dimensional embedding space instead of optimizing directly the full set of parameters.

\section{Related Work}

\textbf{Representations.}
Representation learning concerns the ability of a machine learning algorithm to transform the information contained in raw data in more accessible form.
A common algorithm is the autoencoder \cite{Hinton504}, a self-supervised solution where the representation is learnt by constraining the output to reconstruct the input. 
Our architecture is inspired by the autoencoder but aim at producing outputs that behave akin to the input (e.g. similar performance on a certain task).
In a recent meta-learning paper, LEO \cite{rusu2018meta}, the embedding of the weights of a single layer of a network is learnt for a few shot learning task.
Task2Vec \cite{achille2019task2vec} learns a task embedding on different visual tasks which enables to predict similarities between them and how well a feature extractor perform on a chosen task.
Differently from all these works, we focus on learning a fixed-size embedding for diverse network architectures from which it is possible to draw ready-to-use weights for a specific task, even for networks unseen during training.

\textbf{Network Parameters Prediction.}
Many works deploy an auxiliary network to obtain the weights of a target network. 
Hypernetworks \cite{ha2016hypernetworks} trained a small network (the hypernetwork) to predict weights for a large target network on a given task. 
The same technique has been extended and applied in many ways: transforming noise into the weights of a target network (Bayesian setting) \cite{krueger2017bayesian, louizos2017multiplicative}, adapting the weights of a target network to the current situation \cite{bertinetto2016learning, jia2016dynamic},  generating weights corresponding to hyperparameters \cite{lorraine2018stochastic}, focusing on the acceleration of the architecture search problem \cite{brock2017smash}.
Moreover, networks that generate their own weights have been proposed and analyzed \cite{hyun2018self, chang2018neural}. 
While these works share the use of a weights generation module with our work, our novel proposal consists in showing how to learn a fixed-size structured embedding for different architectures and navigate through this space to obtain new weights for these kinds of architectures as well as for architectures not provided as training examples.

\textbf{Weight-sharing NAS.}
In Weight-sharing NAS, optimal architecture search occurs over the space defined by the subnets of a large network, the supernet. Commonly, subnets share weights with the supernet and they are available as ready-to-use networks after training.
OFA \cite{cai2019once} starts by training the entire supernet and progresses considering subnets of reduced size. After training, desired subnets are selected with an evolutionary algorithm.
In NAT \cite{lu2020neural}, many conflicting objectives are considered, training only the weights of promising subnets for every objective. 
While these works deal with obtaining ready-to-use networks that obey to desired characteristics, we focus on the embeddability of deep models in a latent space organized according to features of interest and on the possibility of explore such latent space by interpolation or optimization.

\section{Method}
\textbf{Framework.}
In the following, we will use \emph{architecture} to denote the structure of a deep learning model (i.e. number and kind of layers, etc.)
and \emph{instance} for an architecture featuring specific parameter values. Of course, given one architecture there can be many instances with different parameter values.

Our framework, dubbed \framework{} and shown in Fig. \ref{fig:framework}, is able to encode trained instances of different architectures into a fixed-size encoding and to decode this embedding into new instances that behave like the input ones. The parameters of each instance presented in input to our framework are stored by a simple algorithm into a \prepr{} (parameters representation), a 2D matrix whose rows are filled one after the other with the sequence of the unrolled parameters. Further details on the \prepr{} generation algorithm are provided in the Suppl. Material.

\framework{} encoder takes as input the \prepr{} of an instance and produces a small fixed-size embedding, applying first horizontal and then vertical convolutions, alongside with max-pooling. It is worth pointing out that the encoder is designed to produce embeddings of the same size for any input \prepr{} dimension.

The embedding from the encoder is then processed by \framework{} decoder, whose basic block first applies convolutions to increase the depth of the input and then reshapes the intermediate output to grow along spatial dimensions at the cost of depth. Once the required \prepr{}  resolution has been reached, an independent linear scaling is applied to every element of the predicted \prepr{}, with weights and biases learnt during the training. We found that this is needed, in particular, for very deep models, probably because convolutions struggle to predict parameters that are close in the \prepr{} but that belong to distant layers of the target architecture.
The values of the predicted \prepr{} are loaded into a ready-to-use instance.

The building blocks of the encoder and decoder are  specified in more details in Fig. \ref{fig:framework}.
In the remainder of the paper, we will use the term \emph{target} instance to refer to the one in input to \framework{} and the term \emph{predicted} instance to refer to that instantiated with values from the predicted \prepr{}.

\textbf{Single-Architecture Setting.}
In the Single-Architecture setting, \framework{} is used to learn an embedding space for the parameters of multiple instances of a single architecture.
The first scenario that we consider deals with instances that exhibit different \textit{performance} in solving the same task, such as image classification. Thus, during \framework{} training, the objective is to learn how to predict weights that match the performance of the target instances. Akin to common practice in Knowledge Distillation \cite{hinton2015distilling}, this can be achieved by minimizing a loss term $\predloss{}$ that represents the discrepancy between the outputs computed by the target instances and those computed by the corresponding predicted instances.
Formally, considering a target instance $N_t$ and training samples $x$ with labels $y$, we denote by $N_p$ the instance predicted by \framework{} when the input instance is $N_t$, and by $t = N_t(x)$ and $p = N_p(x)$ the logits computed by the target and predicted instances, respectively.
We then realize $\predloss{}$ as in \cite{hinton2015distilling}:
\begin{equation}
    \predloss{} = \klloss{}(\softmax(p/T),\;\softmax(t/T)) \cdot T^2
\end{equation}
where $\klloss{}$  denotes the Kullback–Leibler divergence averaged across the samples. As in  \cite{hinton2015distilling}, the softmax functions used in $\predloss{}$ have inputs divided by a temperature term $T$.

A second scenario deals with networks sharing the same architecture that are trained to fit different signals. In particular, recent works \cite{park2019deepsdf, mescheder2019occupancy, sitzmann2020implicit, mildenhall2020nerf} have shown that it is possible to build neural representations of  signals by training MLPs to regress such signals. In this scenario, each instance of the same MLP architecture is trained to represent a different signal. Given one of such instances, the objective of \framework{} is to predict weights capable of regressing the same signal. To achieve this goal, \framework{} can be trained with a loss term that directly compares  the outputs of the predicted instance to those computed by the target one, thereby, also in this case, distilling the knowledge of the target instance into the predicted one.
In this scenario, then, $\predloss{}$ becomes simply:
\begin{equation}
    \predloss{} = MSE(y_p,\;y_t)
\end{equation}
i.e. the Mean Squared Error (MSE) between the outputs from the predicted instance ($y_p$) and those from the target instance ($y_t$) when queried by the same inputs. In particular, in the experiments we consider MLPs trained to regress the Signed Distance Function (SDF) of a 3D shape (e.g., \cite{park2019deepsdf}).

We found that, in both scenarios, using a distillation loss is more effective 
than using a weights reconstruction loss, as the latter would aim just at mimicking on average the weights of the target instances, which we found not implying similar predictions. Furthermore, a distillation loss allows for using each target instance to create many training examples for \framework{} by simply varying the input data.

\textbf{Multi-Architecture Setting.}
In the Multi-Architecture setting, we investigate on how to embed in a common space instances having different architectures. Thus, we consider instances trained to solve an image classification task with the best performances allowed by their architecture. \framework{} is trained to process such instances and to predict weights that reproduce their good performances. In order to ease \framework{} task, we take advantage of the complete Knowledge Distillation described in \cite{hinton2015distilling}. Denoting by $N^*$ a teacher network with good performances in the task at hand and by $t^*$ its logits for a batch of images $x$, we define the loss with respect to it as:
\begin{equation}
    \label{eq:pred_loss_with_reference}
    \predlossmulti{} = \klloss{}(\softmax(p/T),\,\softmax(t^*/T)) \cdot T^2
\end{equation}
Then, as in in \cite{hinton2015distilling}, we introduce an additional term $\taskloss{}$ which, in combination with $\predlossmulti{}$, defines the complete $\kdloss{}$:
\begin{align}
    \taskloss{} &= \celoss{}(\softmax(p),\;y)\\
    \label{eq:kd_loss}
    \kdloss{} &= \alpha \cdot \predlossmulti{} + (1 - \alpha) \cdot \taskloss{} 
\end{align}
where $\celoss{}$ denotes the Cross Entropy loss averaged across the samples of the batch and $\alpha$ is a hyperparameter used to balance the two terms in $\kdloss{}$.

As far as the possibility of handling different architectures is concerned, we identify each architecture uniquely with a categorical  \classid{}. In this configuration, \framework{} is trained to predict an instance with the same architecture as the target. 
Even if this information is available at training time from the target instance itself, we would also like to explore by means of interpolation or optimization the latent space learnt by \framework{} after having trained it, without feeding input instances to the framework. Thus, we wish to be able to extract the architecture information directly from the embedding.
To achieve this objective, we modify the architecture presented so far by adding a softmax classifier on top of the embedding in order to predict the \classid{} of the target instance (details on this variant of the framework are reported in the Suppl. Material). Consequently, we complement the learning objective introduced in Eq. \ref{eq:kd_loss} with an additional $\classloss{}$ term. Given a target instance $N_t$ and the embedding $e$ produced by \framework{} encoder for it, we denote by $c_t$ the \classid{} associated to the architecture of $N_t$ and by $c_p$ the logits predicted by the architecture classifier from $e$. $\classloss{}$ is then defined as the Cross Entropy loss between the predicted and target \classid{}: 
\begin{align}
    \classloss{} &= \celoss{}(\softmax(c_p),\;c_t).
\end{align}

\begin{figure}  
    \begin{center}
    \includegraphics[width=0.45\textwidth]{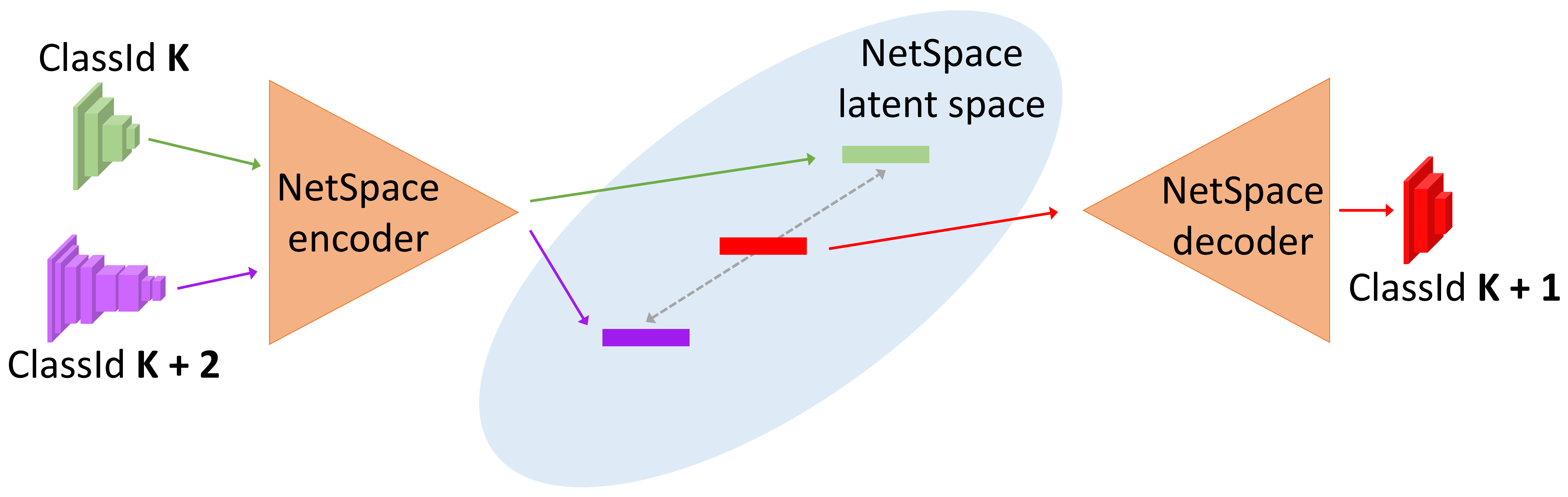}
    \caption{An example of interpolation in the latent space learnt by \framework{} in the Multi-Architecture scenario. \label{fig:interp_loss}}
    \end{center}
\end{figure}

The initial experimental results highlighted that \framework{} was  clustering the latent space according to the architecture \classid{} only. We judge such organization of the embedding space as not satisfactory, as it would allow, perhaps, to sample  new instances within a cluster by proximity or interpolation,  but there would be no simple technique to navigate from one cluster to the others.  Rather, we aim at endowing the embedding space with a structure enabling exploration along meaningful directions, i.e. directions somehow correlated to a specific characteristic, such as number of parameters or performance. Thus, a more amenable organization would consist in  clusters showing up \emph{aligned}, rather than scattered throughout the space,  and possibly  also \emph{sorted} w.r.t. a given characteristic of interest.

Should  such organization of the embedding space be possible, given two \textit{boundary} embeddings (i.e. representing two instances  with the smallest and the largest value of the characteristic of interest), it could be possible to move across the aligned clusters by simply interpolating the boundaries and  obtain along the way representations of ready-to-use instances  with increasing values of the characteristic of interest. 
To further investigate  along this path, we shall consider first that it is possible to assign \classid{}s to a pool of architectures so as to sort them accordingly to a characteristic of interest. For instance, in our experiments, \classid{}s $K$ and $K+1$ will denote two architectures such that the latter has more parameters than the former.

Therefore, we introduce a new loss, denoted as $\interploss{}$ (Interpolation Loss), whose objective is to impose the desired ordered alignment of clusters in the latent space. Given training instances belonging to boundary architectures (i.e. those  with the smallest and largest \classid{}), we first use \framework{} encoder to obtain their embeddings. Then, we interpolate such embeddings according to a given factor $\gamma$, and constrain the interpolated embedding to belong to the architecture whose \classid{} is interpolated between the boundaries according to the same factor $\gamma$. Figure \ref{fig:interp_loss} presents an example of the interpolation procedure described in this paragraph.

Formally, given boundary embeddings $e^A$ and $e^B$ of target instances $N^A_t$ and $N^B_t$ with \classid{}s $c^A$ and $c^B$, we define the interpolated embedding $e^\gamma = (1 - \gamma) \cdot e^A + \gamma \cdot e^B$. Then, considering the logits $c_p^{\gamma}$  predicted by the \classid{} classifier for $e^\gamma$ and the interpolated \classid{} $c^{\gamma}_t = (1 - \gamma) \cdot c^A + \gamma \cdot c^B$, we define $\interplossclass{}$ to impose the consistency of the interpolation factor for \classid{} as:
\begin{align}
    \interplossclass{} &= \celoss{}(\softmax(c_p^\gamma),\;c^{\gamma}_t).
\end{align}

Moreover, considering the instance $N_p^\gamma$ predicted by \framework{} from $e^\gamma$, we denote by $p^\gamma$ the logits predicted by such instance for a batch of images and define $\interplosspred{}$ as:
\begin{align}
    \interplosspredone{} &= \klloss{}(\softmax(p^{\gamma}/T),\,\softmax(t^*/T)) \cdot T^2\\
    \interplosspredtwo{} &= \celoss{}(\softmax(p^\gamma),\;y)\\
    \interplosspred{} &= \alpha \cdot \interplosspredone{} + (1 - \alpha) \cdot \interplosspredtwo{}
\end{align}
with the objective of distilling the teacher network $N^*$ also in the interpolated instances. Finally, we define the total interpolation $\interploss{}$ as:
\begin{align}
    \label{eq:interp_loss}
    \interploss{} &= \interplossclass{} + \interplosspred{}
\end{align}

In our framework, we use $\interploss{}$ with different interpolation factors $\gamma$, whose values are computed according to the number of considered architectures. More precisely, considering $A$ architectures, $\gamma$ can be computed as:
\begin{equation}
    \label{eq:gamma}
    \gamma = \frac{i}{A-1} \quad i \in \{1, 2, ..., A-2\}.
\end{equation}
Given a batch of instances, we compute $\kdloss{}$ and $\classloss{}$ on each of them. Then, we apply $\interploss{}$, with $\gamma$ values obtained from Eq. \ref{eq:gamma}, on all the pairs composed of instances with the minimum and maximum \classid{}. The final loss, thus, is the sum of $\kdloss{}$, $\classloss{}$ for each instance of the batch and $\interploss{}$ for each pair of boundary instances.

\begin{figure*}
    \centering
    \begin{subfigure}[b]{0.24\textwidth}
        \centering
        \includegraphics[width=\textwidth]{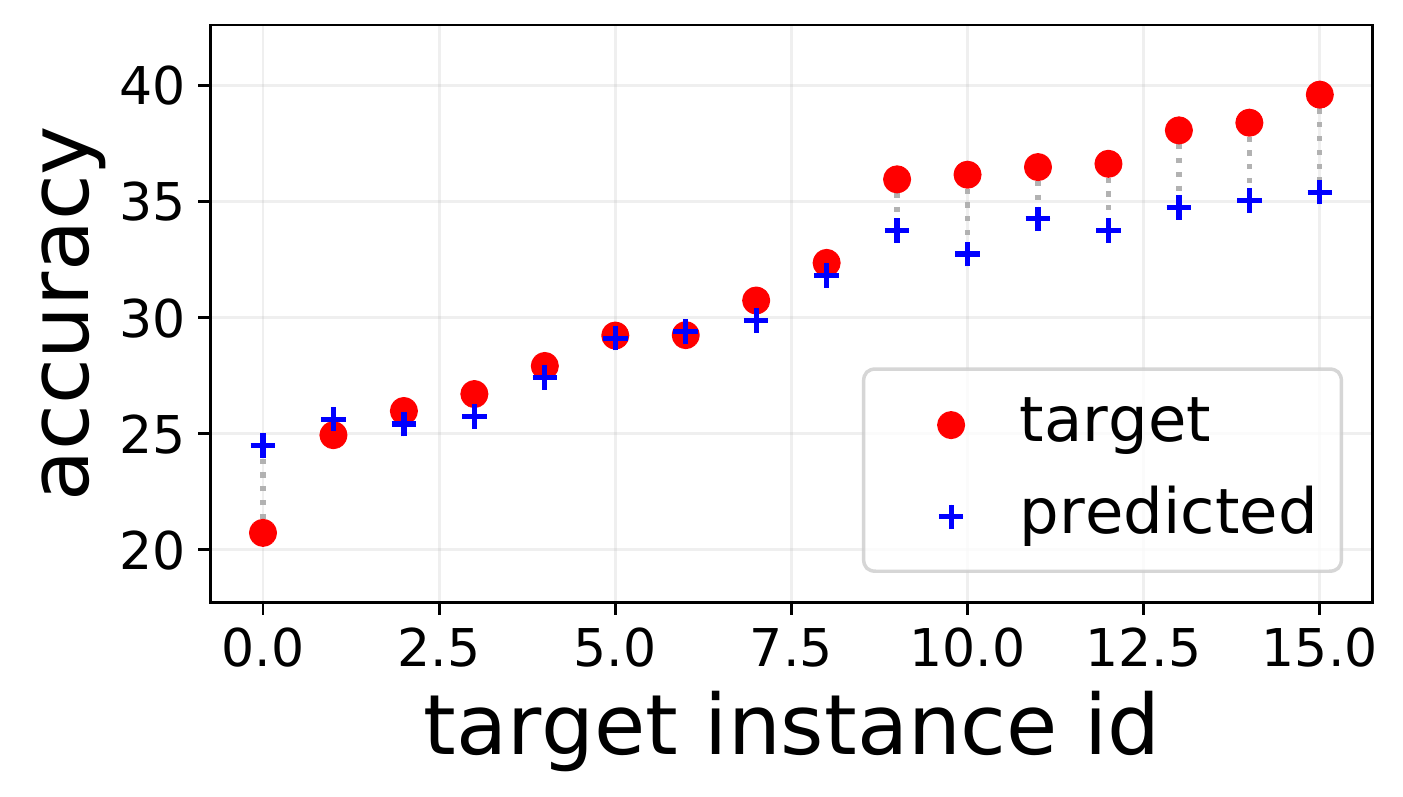}
        \caption{TIN test set}
        \label{fig:single_arch_TIN_test}
    \end{subfigure}
    \begin{subfigure}[b]{0.24\textwidth}
        \centering
        \includegraphics[width=\textwidth]{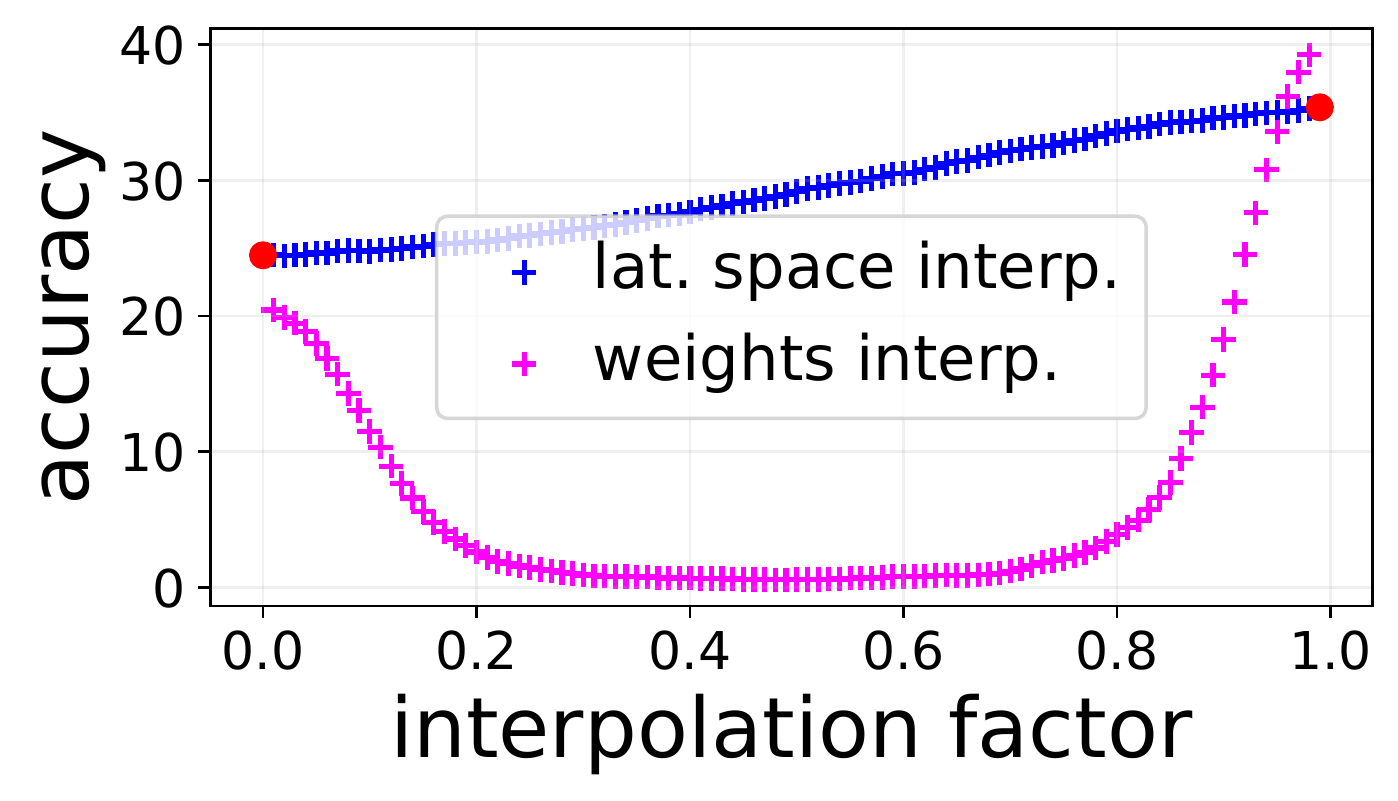}
        \caption{TIN interpolation}
        \label{fig:single_arch_TIN_inter}
    \end{subfigure}
    \begin{subfigure}[b]{0.24\textwidth}
        \centering
        \includegraphics[width=\textwidth]{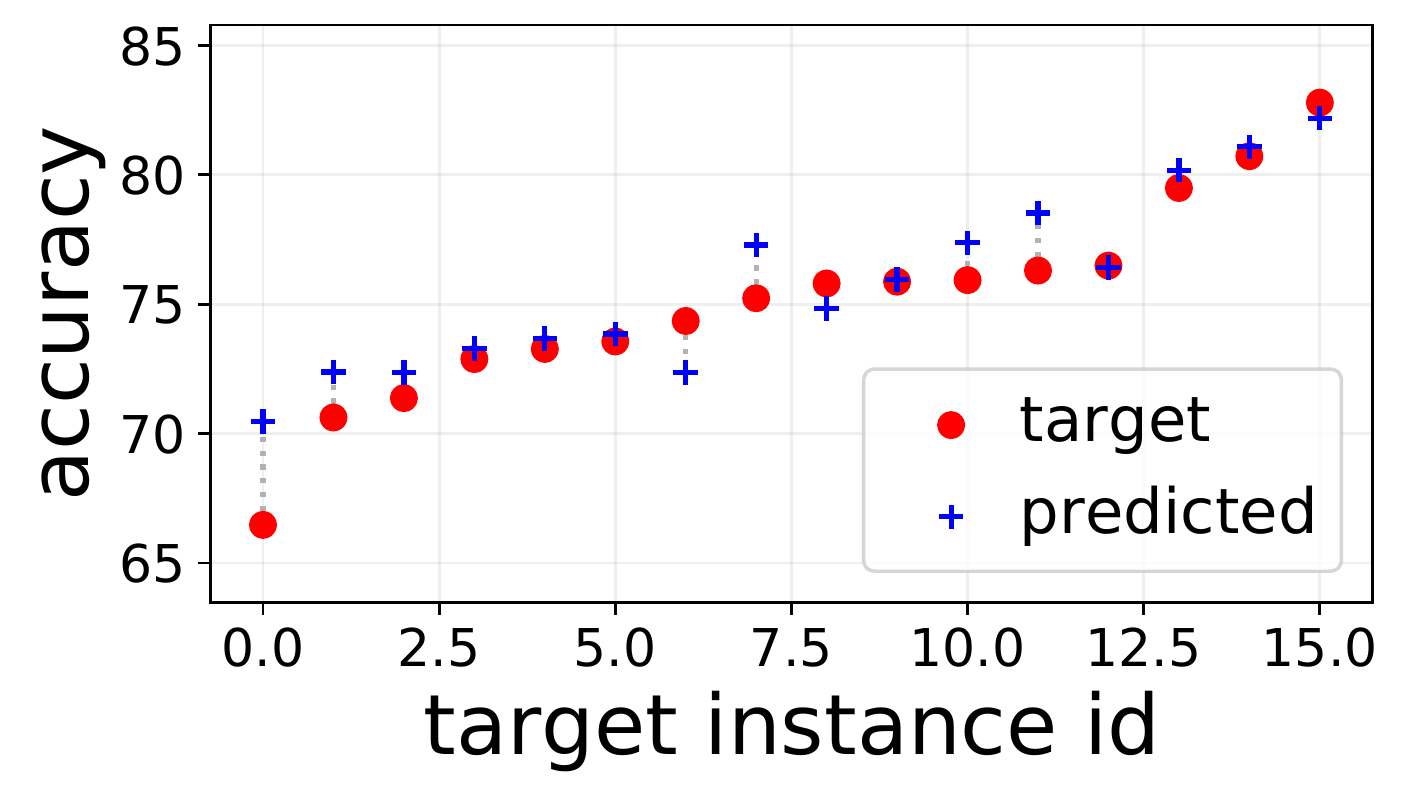}
        \caption{CIFAR10 test set}
        \label{fig:single_arch_CIFAR10_test}
    \end{subfigure}
    \begin{subfigure}[b]{0.24\textwidth}
        \centering
        \includegraphics[width=\textwidth]{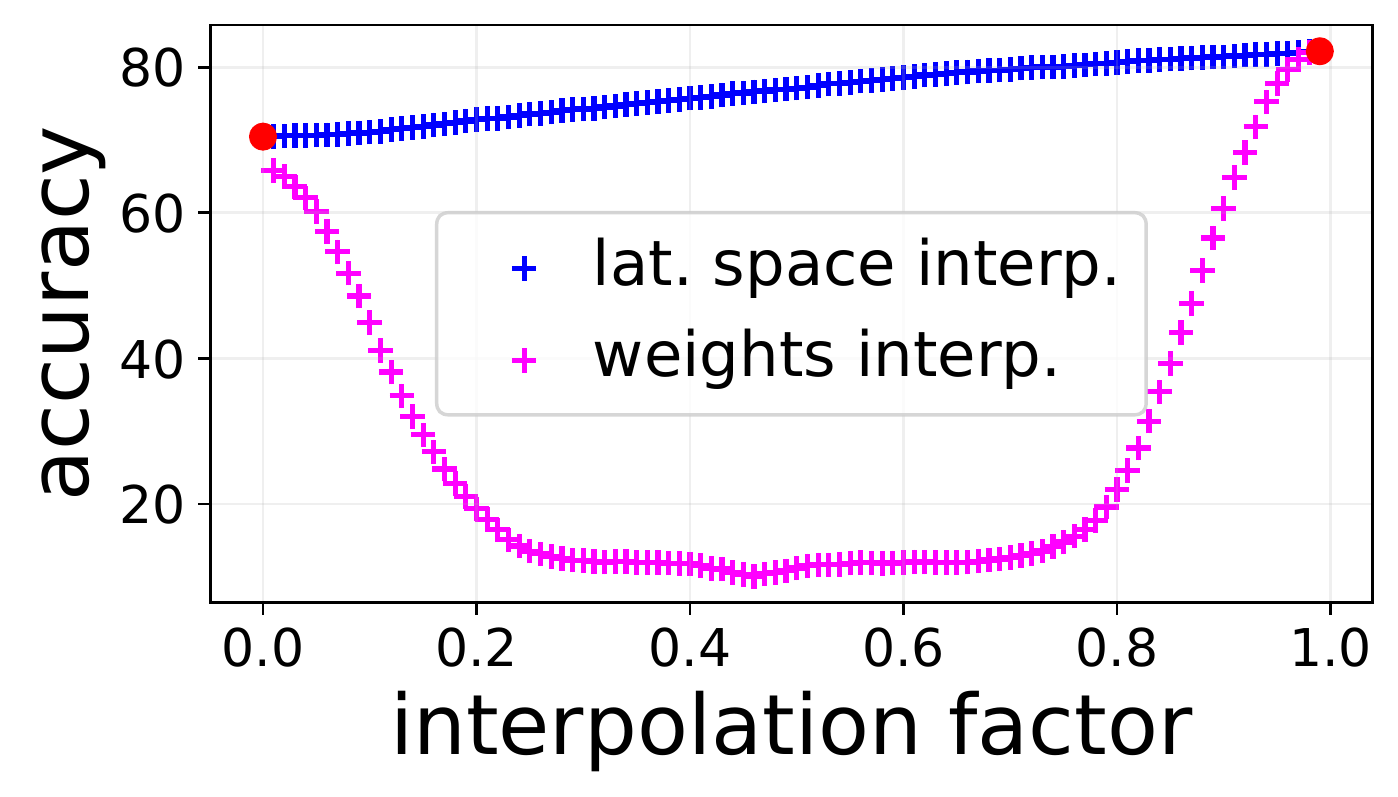}
        \caption{CIFAR10 interpolation}
        \label{fig:single_arch_CIFAR10_inter}
    \end{subfigure}

    \caption{Single-Architecture results for ResNet8. (a) and (c): Accuracy achieved on the test set by target and predicted instances. Target instances are sorted w.r.t. their performances on the test set. (b) and (d): Accuracy achieved on the test set by instances predicted from interpolated embeddings.}
    \label{fig:single_arch}
\end{figure*}

\begin{figure}
    \centering
    \begin{tabular}{ c }
        \includegraphics[width=0.35\textwidth]{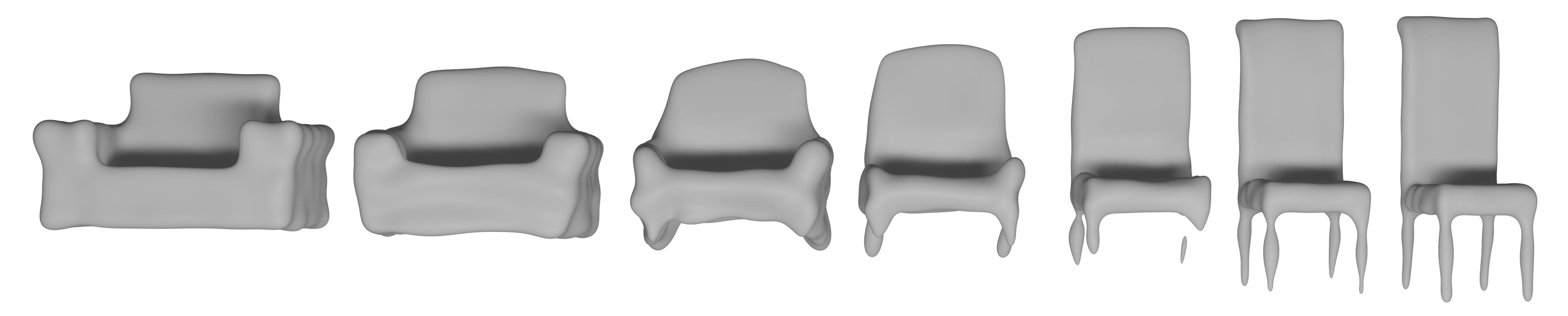}\\
        \includegraphics[width=0.35\textwidth]{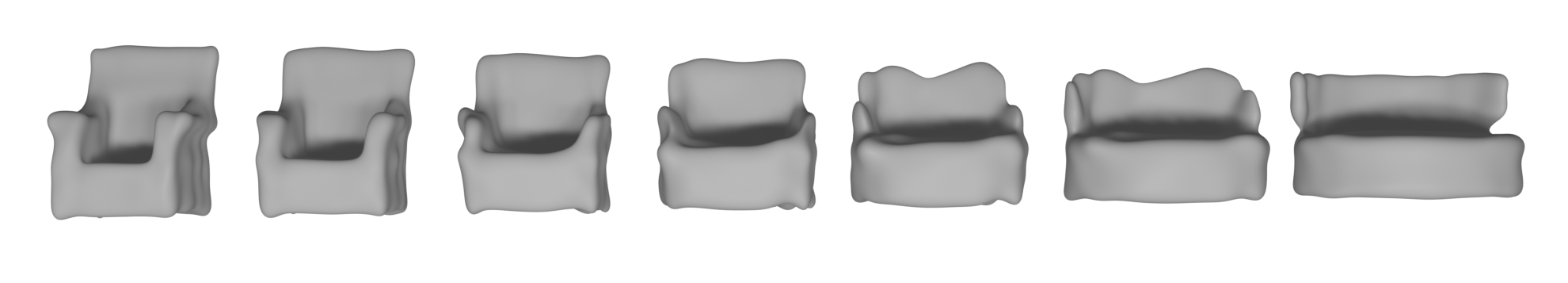}\\
        \hline\\
        \includegraphics[width=0.35\textwidth]{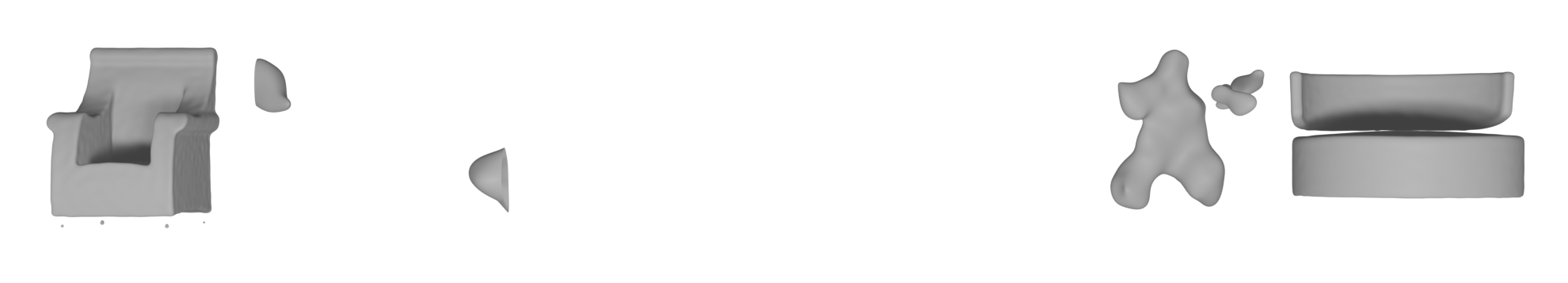}
    \end{tabular}
    \caption{Interpolation of 3D shapes. Top two rows: results obtained by interpolating \framework{} embedding space. Bottom row: the same linear interpolation applied to MLPs weights.}
    \label{fig:shapes}
\end{figure}

\section{Experiments}
We test our framework with networks trained on image classification and 3D SDF regression.

\textbf{Datasets and Architectures.}
For what concerns image classification, we report results on Tiny-ImageNet (TIN) \cite{le2015tiny} and CIFAR-10 \cite{krizhevsky2009learning} datasets.
The target architectures for our experiments are \lenet{}, a slightly modified version of the lightweight CNN introduced in \cite{lecun1998gradient}, \vanillacnn{}, a sequence of standard convolutions followed by a fully connected layer, and two variants of ResNet \cite{he2016deep}, namely ResNet8, and ResNet32. 
As far as 3D SDF regression is concerned, we consider MLPs trained to overfit a selection of $\sim1000$ \textit{chairs} from the Shapenet dataset \cite{chang2015shapenet}. Each MLP has a single hidden layer with 256 nodes and uses periodic activation functions as proposed in \cite{sitzmann2020implicit}.
Additional details are available in the Suppl. Material.

\textbf{Single-Architecture Image Classification.}
As a first experiment, we test \framework{} in the Single-Architecture setting with the image classification task on CIFAR-10 and TIN. We create a dataset of 132 randomly initialized ResNet8 instances, training them for a different numbers of epochs, to collect instances with different performances. Then we randomly select 100 instances for training, 16 for validation, and 16 for testing.
Fig. \ref{fig:single_arch_TIN_test} and \ref{fig:single_arch_CIFAR10_test} compare the accuracy achieved on TIN and CIFAR-10 test sets by target and predicted instances. The target instances belong to the test sets and were never seen by \framework{} at training time. We can see that, beside few outliers, our framework is effective in predicting new instances that emulate the behavior of the target ones, both on CIFAR10 and on the larger and more varied TIN. It is remarkable that \framework{} is able to reconstruct an instance which follows the input one in terms of performance in spite of the huge compression it introduces. Indeed, the embedding size is a fraction of the number of parameters of the instances it can reconstruct, e.g. it is 4096 for TIN, only $\sim3.18\%$ of the parameters of ResNet8. The key information about the behaviour of a neural net seems to live in a low dimensional space. Indeed, as shown by the visualization of PReps provided in the Suppl. Material, the predicted instance is very different from the target one: \framework{} captures the essential information to reproduce the behaviour of the target network, it does not merely learn to reproduce it.

After training, \framework{} has learnt to map target instances to fixed-sized embeddings.
Thus, we can use \framework{} frozen encoder to obtain the embeddings of two anchor instances and linearly interpolate between them in order to study the representations laying in the space between the two anchors. 
To this aim, we decode every interpolated embedding to generate a new instance with \framework{} frozen decoder and compute the accuracy of this ready-to-use instances on the images in the test sets. As a baseline, we consider the possibility of interpolating directly the weights of the anchor instances. Results are reported in Fig. \ref{fig:single_arch_TIN_inter} and \ref{fig:single_arch_CIFAR10_inter} for TIN and CIFAR-10, respectively. 
Interestingly, along the multi-dimensional line connecting the anchor embeddings we find representations corresponding to instances whose performances grow almost linearly with the interpolation factor, while interpolating directly the weights of the anchors yields random performances almost everywhere. This result suggests that the embedding space learnt by our framework can be organized to have meaningful dimensions, that are not exhibited in the instances weights space. In fact, the loss function used in the Single-Architecture training concerns performance and our framework learns \emph{naturally} a latent space that, at least locally, can be explored along a direction strictly correlated with performance.

\textbf{Single-Architecture SDF regression.}
As a second Single-Architecture experiment, we train \framework{} to learn a latent space of MLPs that represent implicitly the SDF of chairs from the ShapeNet dataset. We train our framework on a dataset of $\sim1000$ MLPs: each of them has been trained to overfit a different 3D shape, starting from a different random initialization. The goal of this experiment is to assess if \framework{} is capable of learning a meaningful embedding of 3D shapes, which can then be explored by linear interpolation. 
Thus, after training \framework{}, we obtain two anchor embeddings by processing two input MLPs with \framework{} frozen encoder. Then, we obtain new embeddings by interpolating the anchors and we predict new MLPs with \framework{} frozen decoder. The results of this experiment are reported in Fig. \ref{fig:shapes}. The top two rows show interpolation results obtained from \framework{} latent space, while the bottom row presents results obtained by interpolating directly the weights of the anchor MLPs. We can notice that direct interpolation in the MLPs weights space yields catastrophic failures, while \framework{} embedding space enables smooth interpolations between the boundary shapes. This shows its ability to distill the core content of a trained model into a small-size embedding abstracting from the specific values of weights, and also its flexibility: when the loss concerns fitting of shapes, the latent space of models \emph{naturally} organizes to have dimensions correlated with shape.


\begin{figure*}
    \centering
    \begin{tabular}{ c c c }
        \includegraphics[width=0.25\textwidth]{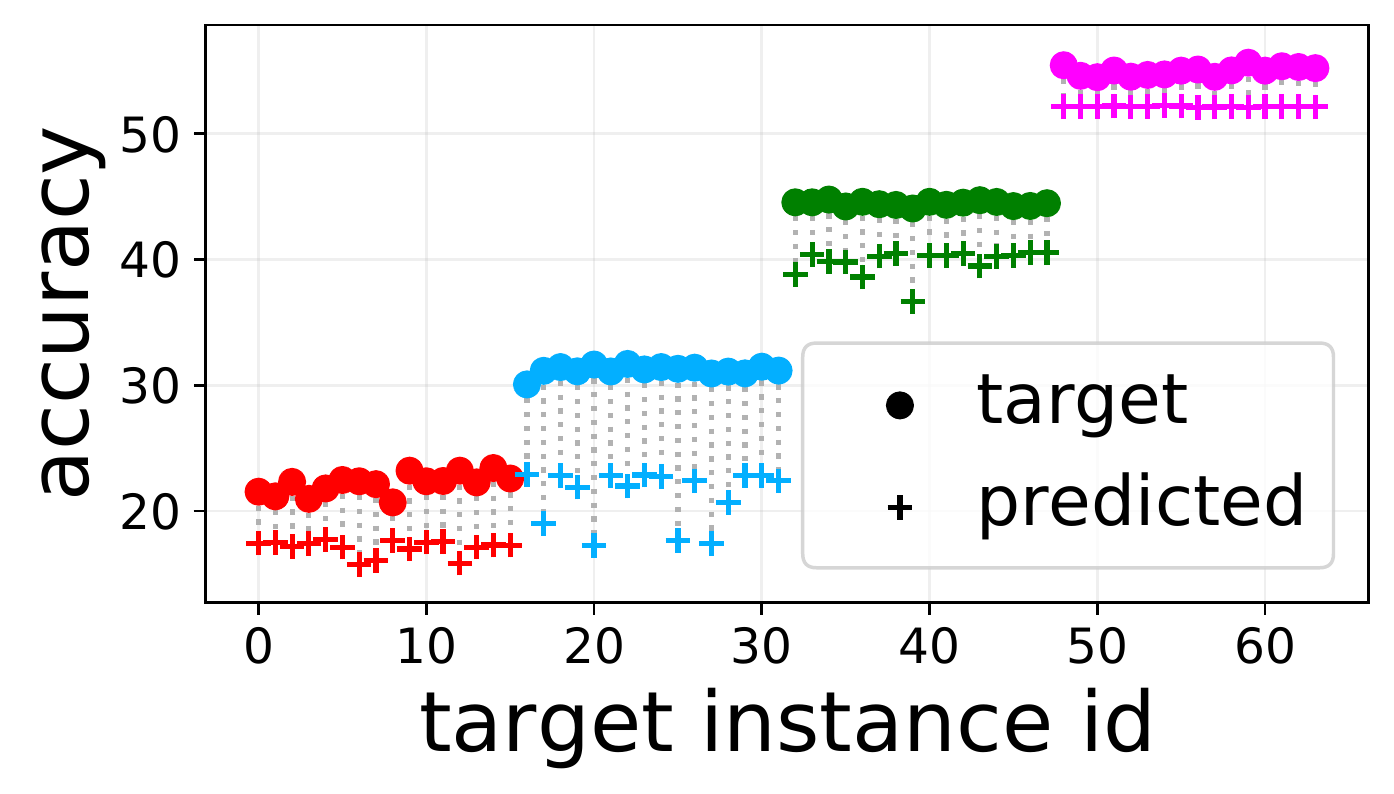} & \includegraphics[width=0.25\textwidth]{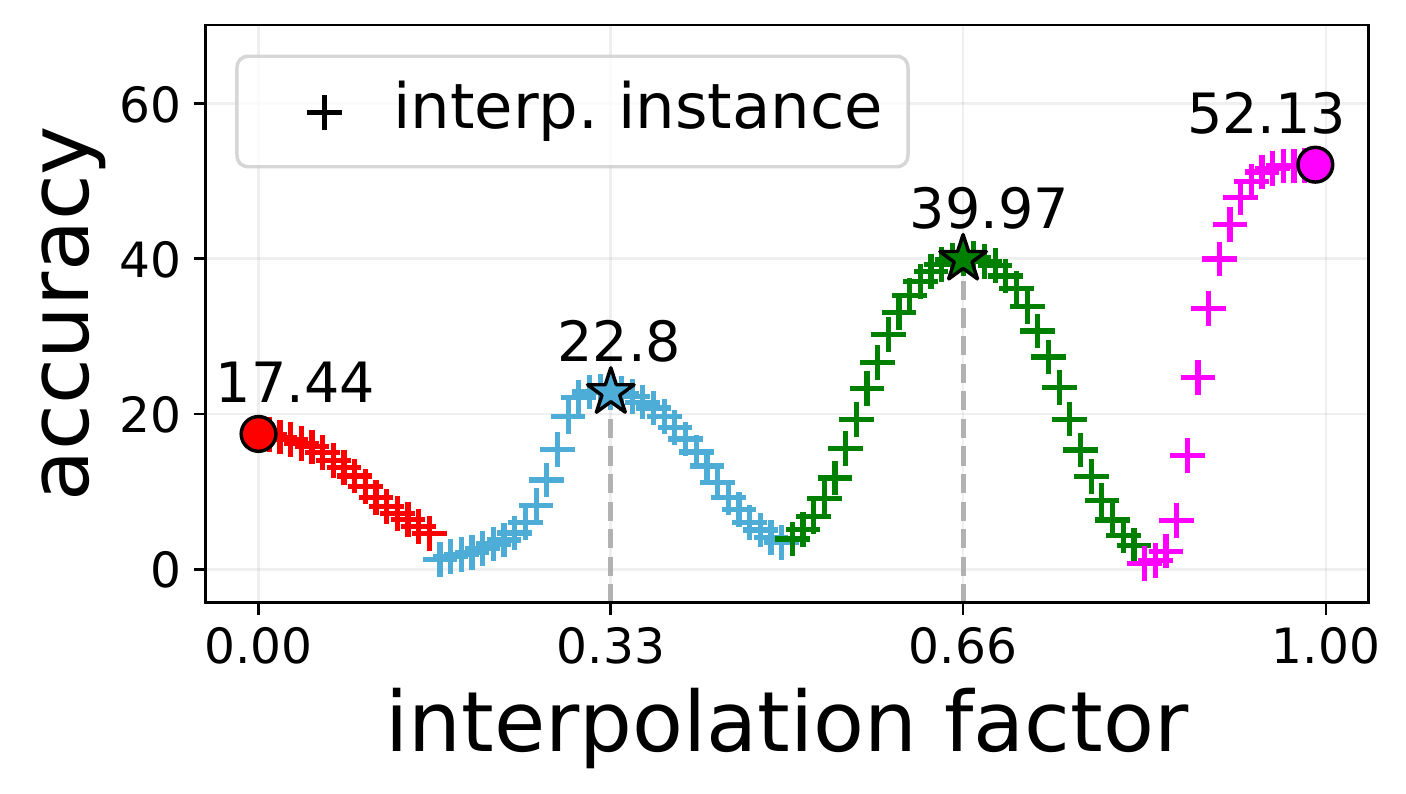} & \includegraphics[width=0.25\textwidth]{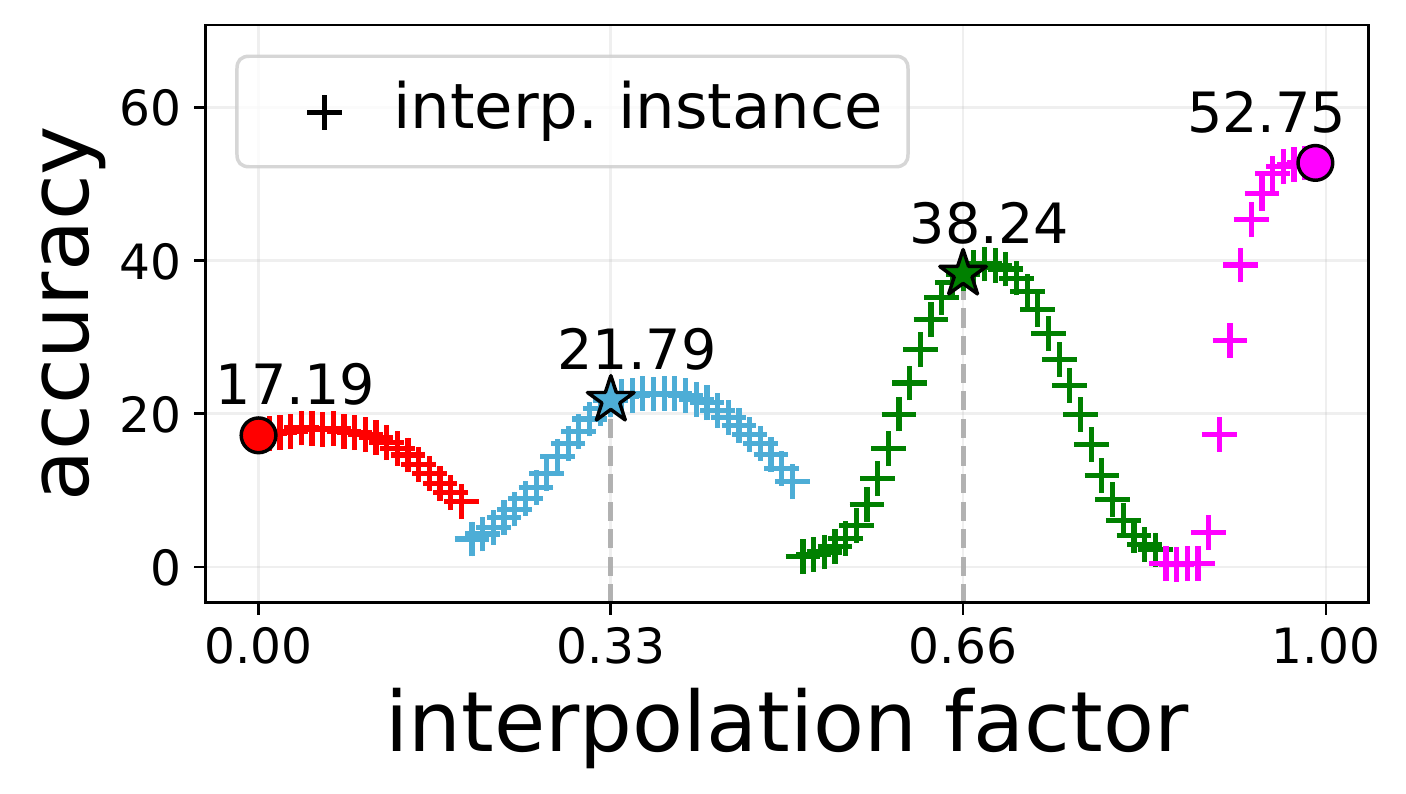}\\
        \includegraphics[width=0.25\textwidth]{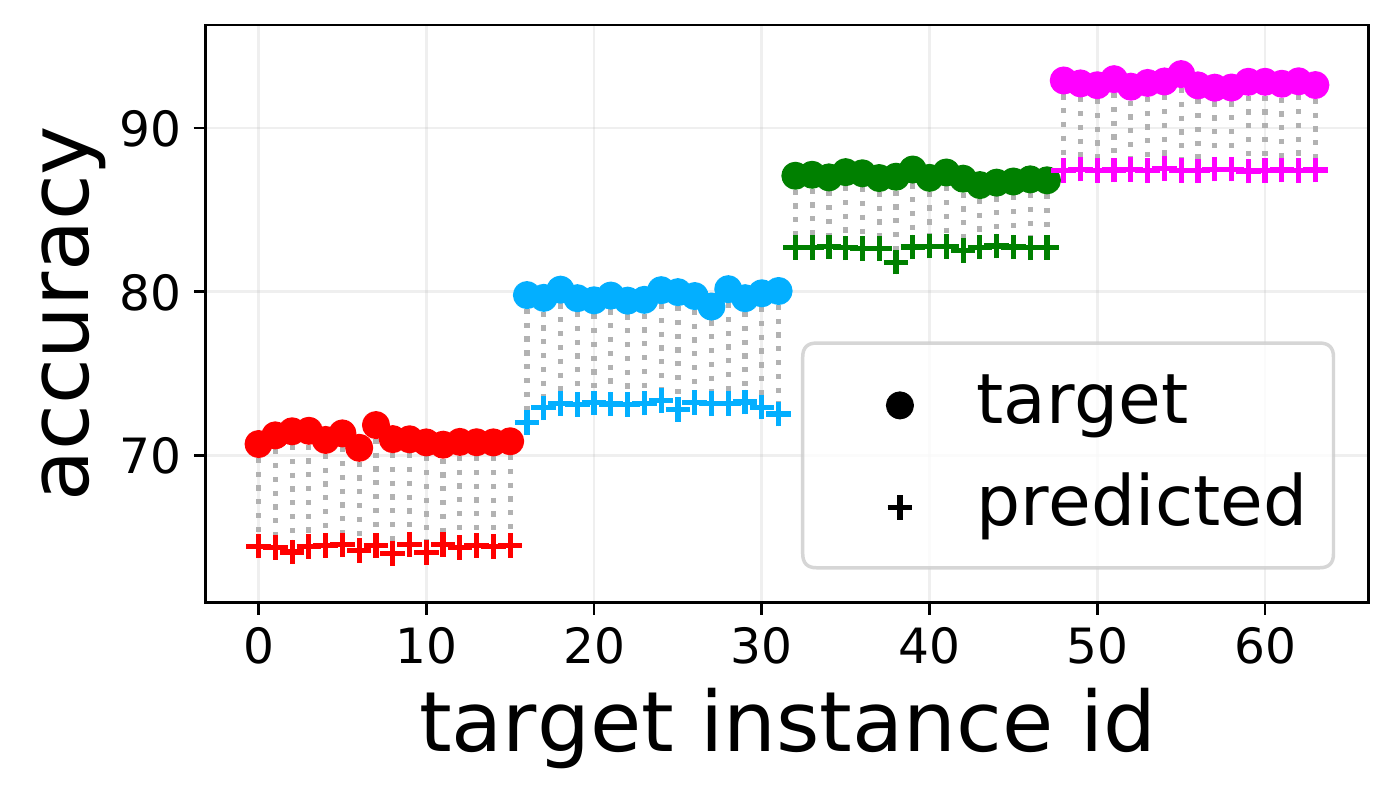} & \includegraphics[width=0.25\textwidth]{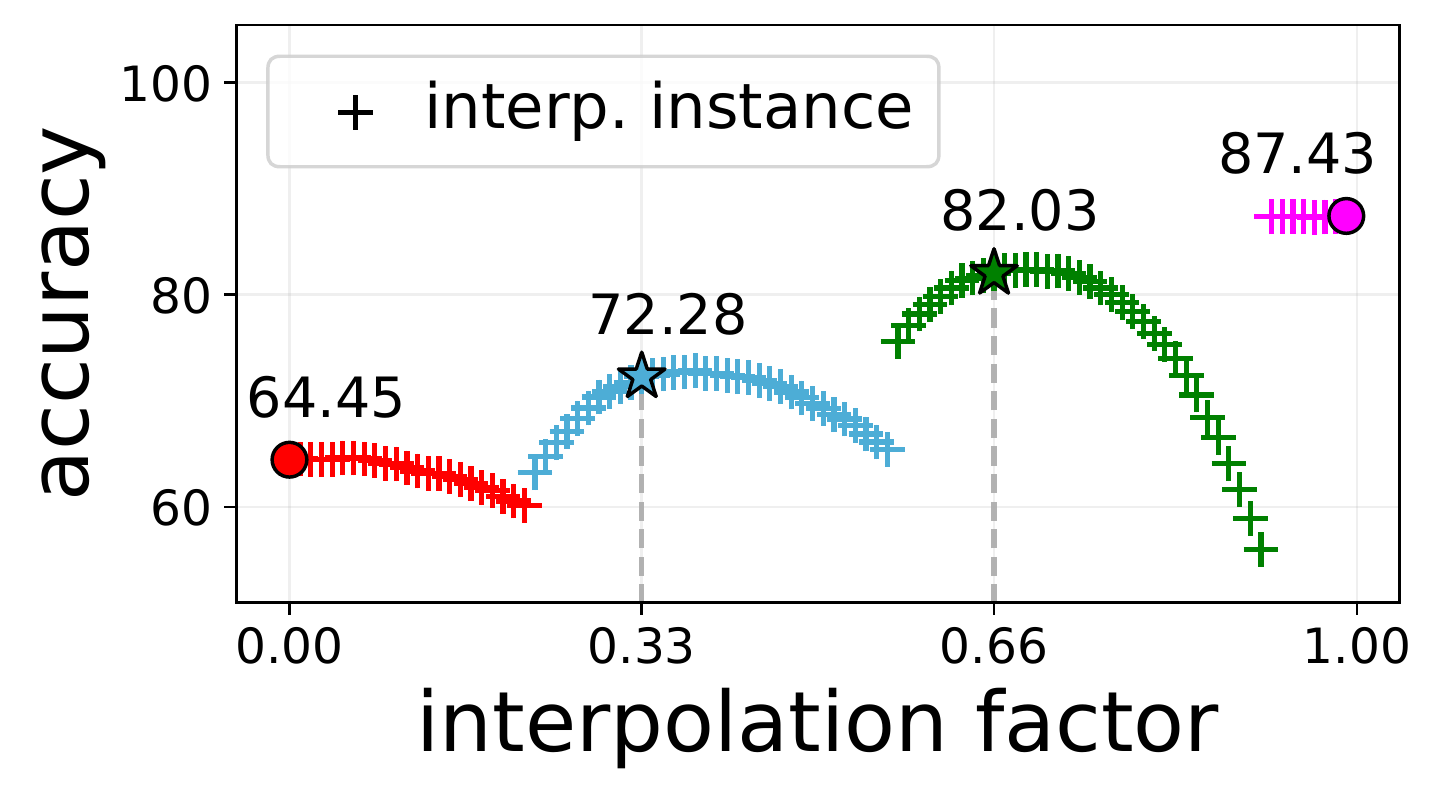} & \includegraphics[width=0.25\textwidth]{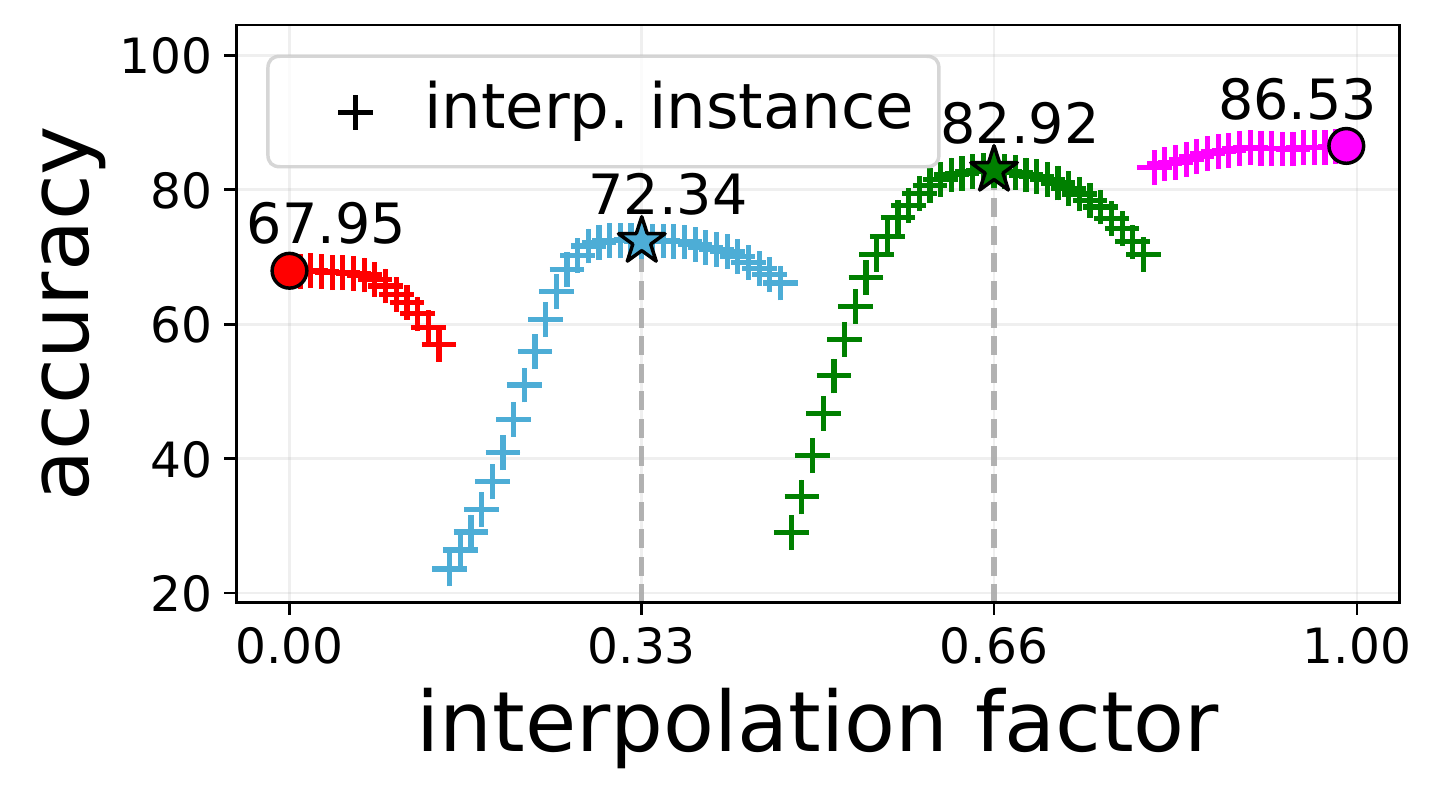}
    \end{tabular}
    \centering
    \caption{Results of the Multi-Architecture setting on TIN (top) and CIFAR10 (bottom). Left: target instances from test set, instances are sorted w.r.t. their ClassId. Center: interpolation with all architectures available at training time. Right: interpolation with only variants of \lenet{} and ResNet32 seen at training time. In all figures, color represent architecture: red-\lenet{}, blue-\vanillacnn{}, green-ResNet8, fuchsia-ResNet32.
    Circles correspond to interpolation boundaries. Stars denote instances obtained with the discrete interpolation factors used in $\interploss{}$.}
    \label{fig:multi_arch}
\end{figure*}

\textbf{Multi-Architecture.}
In the Multi-Architecture setting we train \framework{} to embed four architectures: \lenet{}, \vanillacnn{}, ResNet8 and ResNet32.
To build the dataset, we train many randomly initialized instances for each architecture, collecting multiple instances with good performances (100 for training, 16 for validation and 16 for testing). We collect in total 400 instances for training, 64 for validation and 64 for testing. We adopt a ResNet56 with high performance as the teacher network in Eq. \ref{eq:pred_loss_with_reference} and \ref{eq:interp_loss} and set $\alpha$ to 0.9  in Eq. \ref{eq:kd_loss}.
We observe that supervision is not the same for different architectures: instances with lower performances receive a stronger signal from the $\kdloss{}$ and $\interploss{}$ provides additional supervision for non-boundary architectures.
%
We alleviate this issue modifing the training set so as to include a different number of instances for each architecture : we include 60 \lenet{}, 50 \vanillacnn{}, 60 ResNet8 and 100 ResNet32 instances.
Fig. \ref{fig:multi_arch} left shows the results of this experiment: \framework{} successfully embeds instances of multiple architectures in highly compressed representations with all the information needed to a) predict correctly the architecture of the target instance and b) reconstruct an instance of such architecture whose behavior mimics that of the target one.

\textbf{Multi-Architecture Embedding Interpolation.}
As we defined the \classid{}s of architectures according to their increasing number of parameters, we expect their latent representations to be sorted w.r.t. this characteristic thanks to our interpolation loss $\interploss{}$.
In this experiment, we explore the latent space by observing the classification accuracy achieved by instances obtained when interpolating one embedding of \lenet{} and one of ResNet32, while moving with smaller steps than those defined in Eq. \ref{eq:gamma}. Notably, as shown in Fig. \ref{fig:multi_arch} center, \framework{} learns an embedding space where architectures vary according to their number of parameters along an hyper-line. Moreover, it organized the space to place best performing embeddings for every class around the positions on which $\interploss{}$ was computed.

\textbf{Sampling of Unseen Architectures.}
We perform a new Multi-Architecture experiment by using a training set composed only of \lenet{} and ResNet32 instances. We collect 40 instances for \lenet{} and 80 for ResNet32, with the same balancing strategy discussed above. By not showing to \framework{} encoder any instance of \vanillacnn{} and ResNet8, we deny it the possibility to learn directly a portion of the embedding space dedicated to them. However, $\interploss{}$ shapes it indirectly: for instance, given two embeddings $e_{lenet}$ and $e_{r32}$ of, respectively, a \lenet{} and ResNet32, it forces the embedding $e^{\gamma}=0.33 \cdot e_{lenet} + 0.66 \cdot e_{r32}$ to represent an instance of ResNet8 (unseen during training).
After training, we perform an interpolation experiment and report results in Fig. \ref{fig:multi_arch} right: we find that the latent space learnt by \framework{} trained on a reduced set of architectures exhibits the same properties as the space learnt by training with all of them, allowing to draw by interpolation instances with good performance of the \emph{unseen architectures} \vanillacnn{} and ResNet8.

\begin{table}
    \caption{Accuracy on TIN test set achieved with LSO.}
    \label{tab:latent_opt}
    \centering
    \scalebox{0.73}{
    \begin{tabular}{cc|cccc}
        \toprule
        & \bf Single-Architecture & \multicolumn{4}{c}{\bf Multi-Architecture}\\
        \midrule
        & ResNet8 & \lenet{} & \vanillacnn{} & ResNet8 & ResNet32\\
        \midrule
        initial & 25.73\% & 17.44\% & 22.89\% & 38.81\% & 52.17\%\\
        optimized & 35.72\% & 18.55\% & 24.17\% & 42.07\% & 53.13\%\\
        \bottomrule
    \end{tabular}}
\end{table}

\textbf{Latent Space Optimization.}
Here we investigate the navigation of \framework{} embedding by latent space optimization (LSO).
In order to do so, we take \framework{} encoder and decoder obtained by a Single-Architecture Image classification experiment, freezing their parameters. Then, we obtain an initial latent code by embedding one ResNet8 instance from the test set with the frozen encoder. We then use the frozen decoder to perform an iterative optimization of the initial embedding. In each step of the optimization, the embedding is processed by the frozen decoder, which predicts a \prepr{} that is loaded into a ResNet8 instance.
The resulting network is then used to produce predictions on a batch of training images from TIN. We apply $\kdloss{}$ on these predictions using a ResNet56 as teacher network, to guide \framework{} in the search of a high performing instance in the learnt latent space. As the decoder is frozen, we compute the gradient of the loss  w.r.t. the embedding, so as to explore the latent space by gradient descent.  
Furthermore, we perform a similar experiment starting from \framework{} encoder and decoder taken from a Multi-Architecture training experiment. Also in this case, we use the frozen encoder to obtain initial embeddings from four instances belonging to different architectures. Then, we process each embeddings with the frozen decoder, which predicts a \prepr{} and a \classid{}. We use the predicted \classid{} to select the architecture where the predicted \prepr{} is loaded, building an instance which we use to produce predictions on a batch of training images from TIN. In addition to $\kdloss{}{}$, in this case we apply also $\classloss{}$ on the predicted \classid{}, to guide the embedding towards the area of the latent space that corresponds to the desired class.
In Tab. \ref{tab:latent_opt} we report the performances obtained by the optimizations (second row), together with the performances of the instances used to obtain the initial embeddings (first row). Remarkably, the results show that it is actually possible to improve the performances of the input instances by exploring the \framework{} embedding space via latent space optimization.

\section{Conclusions and Future Work}
\label{sec:conclusions}
\framework{} introduces a framework to learn the latent space of deep models. We have shown 
that the embedding space learnt by \framework{} can be organized according to meaningful traits and ready-to-use instances with predictable properties can be obtained by means of linear interpolation or latent space optimization. Furthermore, our experiments provide 
evidence that fixed-size embeddings can represent effectively instances of several different architectures.
The main limitation of our work concerns lack of investigation on the potential applications of our findings dealing with embeddability of deep models.
Yet, we deem it worth communicating these findings to the community as we believe they are non-obvious and valuable on their own, and may foster further research as well as identification of the most fruitful outcomes toward applications. 

\bibliography{paper}
\bibliographystyle{acm}

\newpage\phantom{Supplementary}
\multido{\i=1+1}{7}{


\section*{Supplementary material}

# Supplementary Material for Learning the Space of Deep Models

Gianluca Berardi*, Luca De Luigi*, Samuele Salti, Luigi Di Stefano

Department of Computer Science and Engineering (DISI), University of Bologna, Italy

{gianluca.berardi3, luca.deluigi4, samuele.salti , luigi.distefano}@unibo.it

* Corresponding authors with equal contribution.

## I. CLASSID CLASSIFIER

In Sec. 3 (*Multi-Architecture Setting*) of the main paper we introduce the need to extend NetSpace architecture to the Multi-Architecture setting. In this setting, in fact, we ask our framework to extract the ClassId of the predicted instances from the embeddings. To achieve this goal, we extend the architecture of our framework with a softmax classifier, which takes in input the embeddings generated by NetSpace encoder and is trained to predict the correct ClassId with $\mathcal{L}_{class}$. The classifier is a lightweight neural network, composed of one convolutional layer and one fully connected layer, interleaved by the LeakyReLU activation function. The outputs of the classifier are transformed into probabilities by the softmax function. Fig. 1 presents an overview of the version of NetSpace used in the Multi-Architecture case, including the ClassId classifier.

## II. PARAMETERS REPRESENTATION (PREP)

A PRep is a 2D tensor where we store all the parameters of an instance of a neural network by means of a simple algorithm exemplified in Fig. 2.

We designed our framework fixing PReps to be rectangular matrices with high width/height ratio. Additionally, to favor easy implementation, the height and the width of a PRep are fixed to be multiple of 4 and 8, respectively. Considering an architecture with $P$ parameters and having fixed the width of the PRep to a number $W$ (divisible by 8), the minimum necessary height of the PRep can be computed as $\lceil \frac{P}{W} \rceil$, adding a padding of $4 - (\lceil \frac{P}{W} \rceil \mod 4)$ rows, if needed, to fulfill the divisibility by 4.

Given an instance of a neural network and a chosen PRep size, the algorithm to produce the instance PRep is straightforward: parameters from all the layers of the instance are copied in the matrix in sequence one row after the other and final zero-padding is added as needed to match the required size.

## III. NETWORK ARCHITECTURES

**Image Classification.** Fig. 3 and 4 present the architecture of the neural networks used in our experiments dealing with image classification, i.e. LeNetLike [6], VanillaCNN and ResNet8/32/56 [2]. Each convolutional layer is presented in the form CONV (in $I$, out $O$, k $K$, s $S$, p $P$), where $I, O, K, S$ and $P$ represent input channels, number of filters, kernel size, stride and padding, respectively. Fully Connected layers, instead, are reported in the form FC (in $I$, out $O$), where $I$ and $O$ stand for input and output units, respectively. Average pooling is shown as Average Pooling (k $K$, s $S$), where $K$ is kernel size and $S$ is stride. Finally, CONV 1×1 represents a convolutional layer with kernel size 1 used to adapt feature maps in residual connections.

**3D SDF Regression.** As far as 3D SDF regression is concerned, we use simple Multilayer Perceptrons (MLPs) composed of a single hidden layer with 256 nodes. Following [8], we use a periodic activation function between the input layer and the hidden layer and between the hidden layer and the output layer. No activation function is applied instead on the final outputs.

## IV. IMAGE CLASSIFICATION: EXPERIMENT DETAILS

**Images Datasets.** To test our general framework in image classification, we make use of the CIFAR-10 [4] and Tiny-ImageNet [5] datasets. CIFAR-10 is composed of 60K 32×32 colour images, 50K for training and 10K for test, categorized in 10 classes. For our experiments we obtain a validation set by splitting the training set in 40K images for training and 10K for validation, while we keep unchanged the test set. Tiny-ImageNet consists of 100K colour images with resolution 64 × 64, categorized in 200 different classes. We split the training set in 80K images for training and 20K images for validation and use the 10K images of the provided validation set for testing. With both datasets, we follow a standard data augmentation regime [9] and use a batch size of 128.

**Nets Datasets.** In the Single-Architecture setting, we train the instances in input to NetSpace with the Adam optimizer [3] and constant learning rate set to 0.0001 for a number of epochs that vary from 1 to 600 epochs, obtaining instances with weights and performances varying smoothly across the training iterations. Then we use 100 instances for training, 16 for validation and 16 for test. In the Multi-Architecture setting, we aim at embedding only instances with high performances. Accordingly, we found it more effective to train models with the SGD optimizer for 300 epochs and setting momentum and weight decay to 0.9 and 5e-4, respectively. We set the initial learning rate to 0.05 and decay it by 0.1 at epochs 150, 180 and 210. As discussed in the main paper in Sec. 4, for the Multi-Architecture setting we define a training dataset with 60 LeNetLike instances, 50 VanillaCNN instances, 60

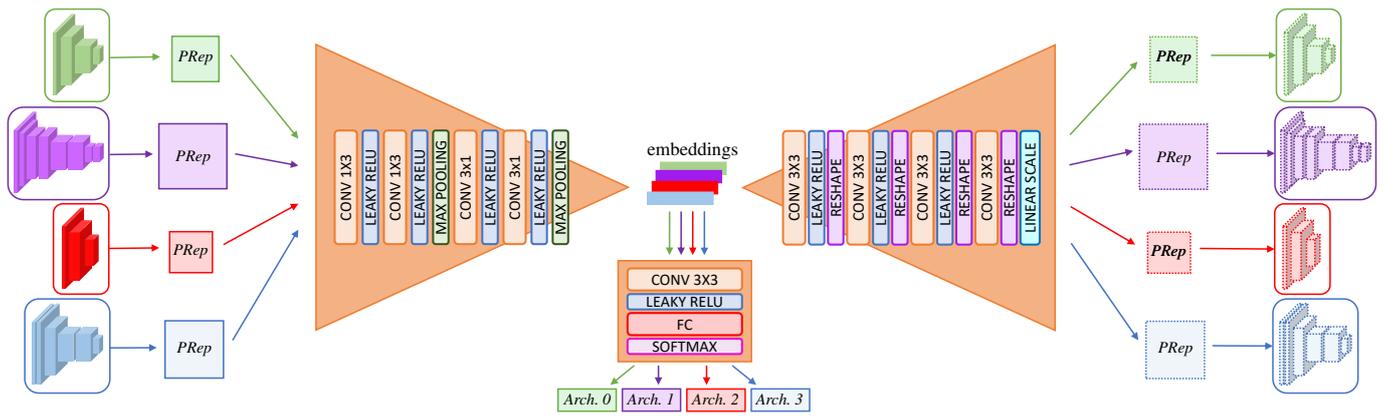

Fig. 1: Overview of NetSpace architecture used in the Multi-Architecture experiments.

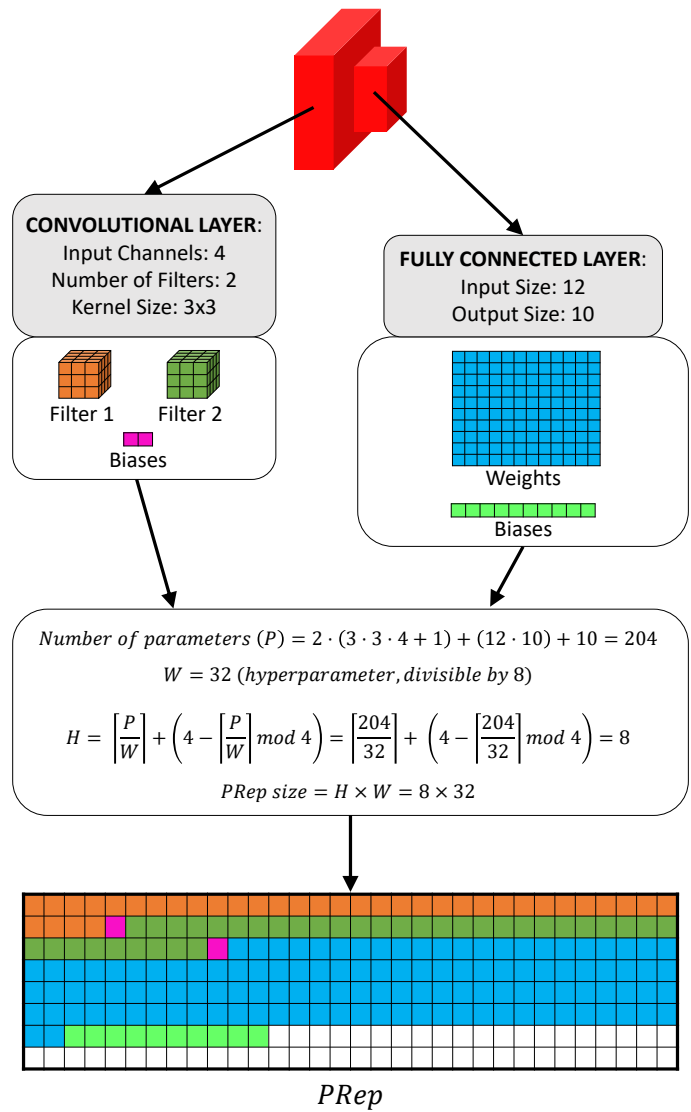

Fig. 2: Algorithm to compute the PRep of a given instance. We consider here a toy architecture made out of one convolutional layer and one fully connected layer and we fix the PRep width to 32. White cells represent padding with constant value 0.

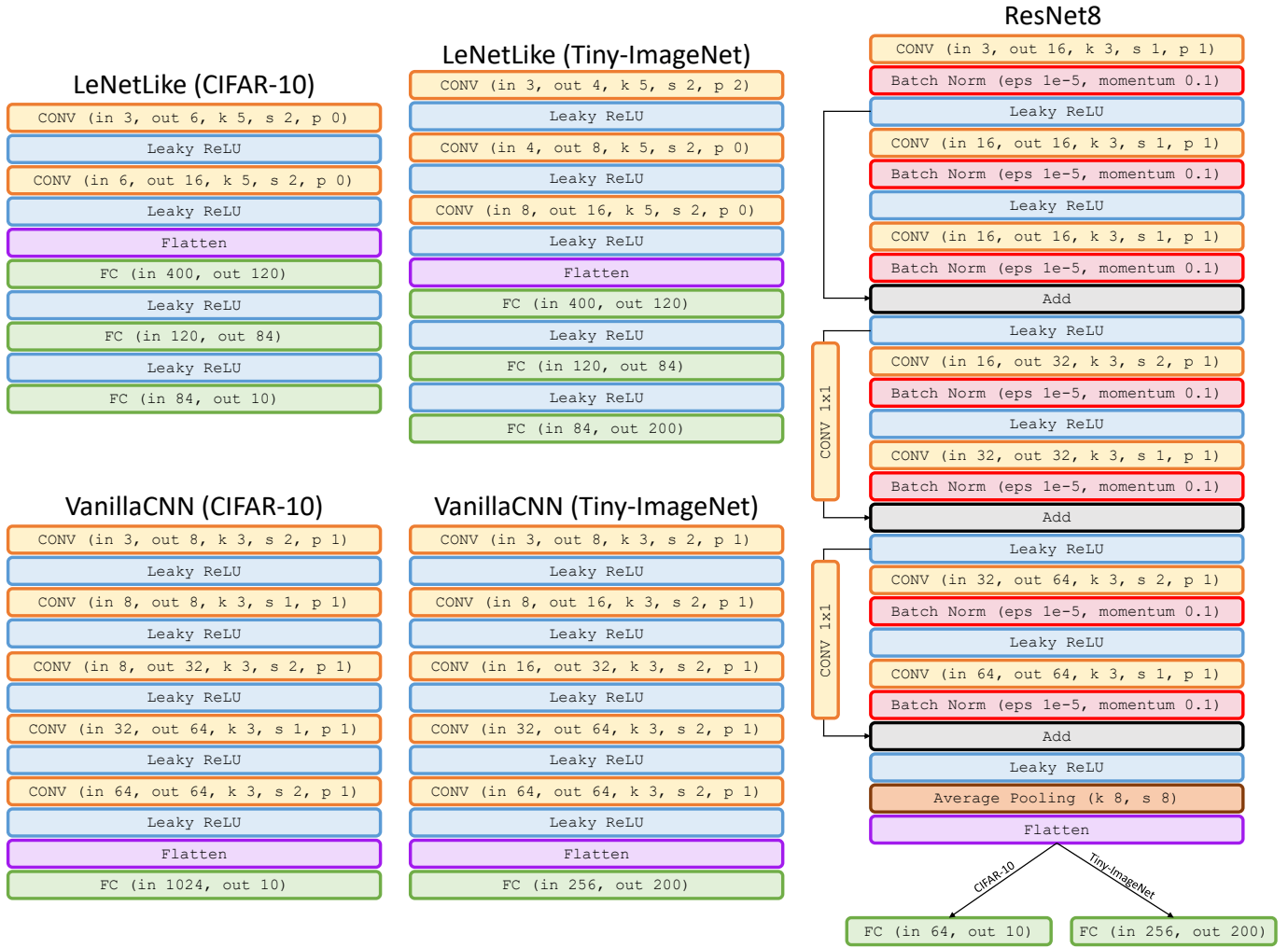

Fig. 3: Architectures used in our experiments dealing with image classification. Top left: LeNetLike, bottom left: VanillaCNN, right: ResNet8.

ResNet8 instances and 100 ResNet32 instances, while for the experiment "Sampling of Unseen Architectures" we use a training set composed of 40 LeNetLike instances and 80 ResNet32 instances. In both Multi-Architecture settings, the validation and test sets are composed of 16 instances of the architectures available during training. In Tab. I we report the accuracy achieved by the trained models on the CIFAR-10 and the Tiny-ImageNet test sets, alongside with the number of parameters of each architecture.

**Framework Training.** To train NetSpace in the Single-Architecture and in the Multi-Architecture settings, we use the Adam optimizer and a learning rate value of 0.0001, we train for around 1K epochs, and then we use the model with the highest performance on the validation set. The temperature term used in $\mathcal{L}_{kd}$ and in $\mathcal{L}^\gamma$ is set to 4, while the size of the meta-batch of instances is set to 8 and to 2 in the Single-Architecture and in the Multi-Architecture settings, respectively. The Latent Space Optimization experiments are conducted by training NetSpace with the Adam optimizer and constant learning rate 0.0001, stopping the training when the accuracy achieved by the optimized network doesn't show any additional improvement.

**Computational Time and Resources.** For our experiments we used several Nvidia RTX 3090 GPUs. Training the networks to populate the datasets requires approximately 600 GPU hours, while training NetSpace with Tiny-ImageNet requires around 84 GPU hours in the Multi-Architecture setting and around 48 GPU hours in the Single-Architecture setting. Training over CIFAR-10, instead, requires less time, with less then 48 and 36 GPU hours respectively in the Multi-Architecture and in the Single-Architecture settings. Finally, the Latent Space Optimization experiments require few GPU hours, but it's possible to observe good improvements over the initial performance already after few minutes of training.

## V. 3D SDF REGRESSION: EXPERIMENT DETAILS

**3D Shape Dataset.** To test NetSpace with networks dealing with 3D SDF regression, we use the ShapeNet dataset [1]. In particular, we use ShapeNetCore, a subset of the full ShapeNet

Fig. 4: Architectures used in our experiments dealing with image classification. Left: ResNet32, right: ResNet56.

TABLE I: Models used in our experiments. We show classification accuracies on the CIFAR-10 (top) and Tiny-ImageNet (bottom) test sets alongside the number of parameters for each architecture. For the Single-Architecture setting (Adam optimizer), we report the accuracy achieved by the best performing network between the trained ones, while for the Multi-Architecture setting (SGD optimizer), we report the average accuracy of the networks that compose the training set.

CIFAR-10

| Net | ClassId | Acc. (Adam) | Acc. (SGD) | # Params |
|---|---|---|---|---|
| LeNetLike | 0 | - | 70.87 % | 62,006 |
| VanillaCNN | 1 | - | 79.77 % | 68,818 |
| ResNet8 | 2 | 82.79 % | 87.13 % | 78,042 |
| ResNet32 | 3 | - | 92.70 % | 466,906 |
| ResNet56 | - | - | 92.85 % | 855,770 |

Tiny-ImageNet

| Net | ClassId | Acc. (Adam) | Acc. (SGD) | # Params |
|---|---|---|---|---|
| LeNetLike | 0 | - | 22.13 % | 79,612 |
| VanillaCNN | 1 | - | 31.32 % | 112,856 |
| ResNet8 | 2 | 40.51 % | 44.38 % | 128,792 |
| ResNet32 | 3 | - | 54.81 % | 517,656 |
| ResNet56 | - | - | 56.93 % | 906,520 |

dataset which covers 55 common object categories with about 51,300 unique 3D models. We conduct our experiments on a small subset of ShapeNetCore, consisting of 1000 3D objects from the *chair* category.

**MLP Dataset.** The MLP dataset used in our experiments contains 1000 MLP. Each of them is obtained by training a randomly initialized MLP to fit the SDF of a single *chair*, whose ground-truth is computed with the code provided with [7]. The fitting procedure consists in 10,000 gradient descent steps: at each step the MLP is queried on 20,000 random 3D coordinates and is asked to regress the SDF value for each of them. Then, the parameters of the MLP are optimized by Adam [3], using as loss function the mean squared error between the predictions and the ground-truth. The learning rate is initially set to 0.0001 and multiplied by 0.9 every 1000 steps.

**Framework Training.** NetSpace is trained on the 1000 MLPs with the protocol described in Sec. 3 of the main paper, using Adam [3] with learning rate set to 0.0001. During training, we evaluate the performance of NetSpace by comparing directly the predictions of the output MLPs with those of the input MLPs: by querying input and output MLPs with the same random coordinates, we can compute the percentage of predictions of the output MLPs that are sufficiently close to the values predicted by the input MLPs. We monitor this metric and stop the training when it reaches the value 0.8. The results reported in the main paper are obtained by training for 4000 epochs.

**Computational Time and Resources.** We adopted Nvidia RTX 3090 GPUs also in the experiments involving SDF regression. The creation of the MLP dataset requires approximately 20 GPU hours, while training NetSpace requires around 48 GPU hours.

## VI. FUSING BATCH NORM AND CONVOLUTIONS

To be able to process architectures including batch norm layers, e.g. ResNet in our experiments, without changing the PRep structure, we decided to fuse batch norm layers with convolutional layers, which is always possible for a trained model since batch norm becomes a frozen affine transformation at test time. Therefore, when processing ResNet instances in our experiments, we first prepared a dataset of instances trained with batch norm to achieve the best performances and then, by the process described below, we transformed such instances into equivalent ones featuring only plain convolutional layers without batch norm.

If we consider a feature map $F$ with shape $C \times H \times W$, at inference time its batch normalized version $\widehat{F}$ is obtained by computing at each spatial location $i, j$:

$$\begin{pmatrix} \widehat{F}_{1,i,j} \\ \widehat{F}_{2,i,j} \\ \vdots \\ \widehat{F}_{C,i,j} \end{pmatrix} = W_{BN} \cdot \begin{pmatrix} F_{1,i,j} \\ F_{2,i,j} \\ \vdots \\ F_{C,i,j} \end{pmatrix} + b_{BN} \quad (1)$$

with

$$W_{BN} = \begin{pmatrix} \frac{\gamma_1}{\sqrt{\widehat{\sigma}_1^2 + \epsilon}} & & & \\ & \frac{\gamma_2}{\sqrt{\widehat{\sigma}_2^2 + \epsilon}} & & \\ & & \ddots & \\ & & & \frac{\gamma_C}{\sqrt{\widehat{\sigma}_C^2 + \epsilon}} \end{pmatrix} \quad (2)$$

$$b_{BN} = \begin{pmatrix} \beta_1 - \gamma_1 \frac{\widehat{\mu}_1}{\sqrt{\widehat{\sigma}_1^2 + \epsilon}} \\ \beta_2 - \gamma_2 \frac{\widehat{\mu}_2}{\sqrt{\widehat{\sigma}_2^2 + \epsilon}} \\ \vdots \\ \beta_C - \gamma_C \frac{\widehat{\mu}_C}{\sqrt{\widehat{\sigma}_C^2 + \epsilon}} \end{pmatrix} \quad (3)$$

where $\widehat{\mu}_c$, $\widehat{\sigma}_c^2$, $\beta_c$ and $\gamma_c$ ($c = 1, 2, \ldots, C$) are respectively the mean, variance and batch norm parameters computed during training for the channel $c$ of the feature map. From this formulation, we can see that batch norm can be implemented as a $1 \times 1$ convolution and therefore, when batch norm comes after another convolution as in ResNet, we can fuse these two convolutions into a single one.

We can express a convolutional layer with kernel size $k$ processing the $C_{prev} \times k \times k$ volume at the spatial location $(i, j)$ of a feature map $F_{prev}$ with $C_{prev}$ channels to produce the feature map $\tilde{F}$ with $C$ output channels as an affine transfomation

$$\tilde{f}_{i,j} = W_{conv} f_{i,j} + b_{conv}, \quad (4)$$

where $W_{conv} \in \mathbb{R}^{C \times (C_{prev} k^2)}$, $b_{conv} \in \mathbb{R}^C$ and $f_{i,j}$ represents the area of size $C_{prev} \times k \times k$ around cell $(i, j)$ reshaped as a $(C_{prev} k^2)$-dimensional vector.

If the batch norm defined by $W_{BN} \in \mathbb{R}^{C \times C}$ and $b_{BN} \in \mathbb{R}^C$ presented in Eq. 2 and 3 comes after such convolutional layer,

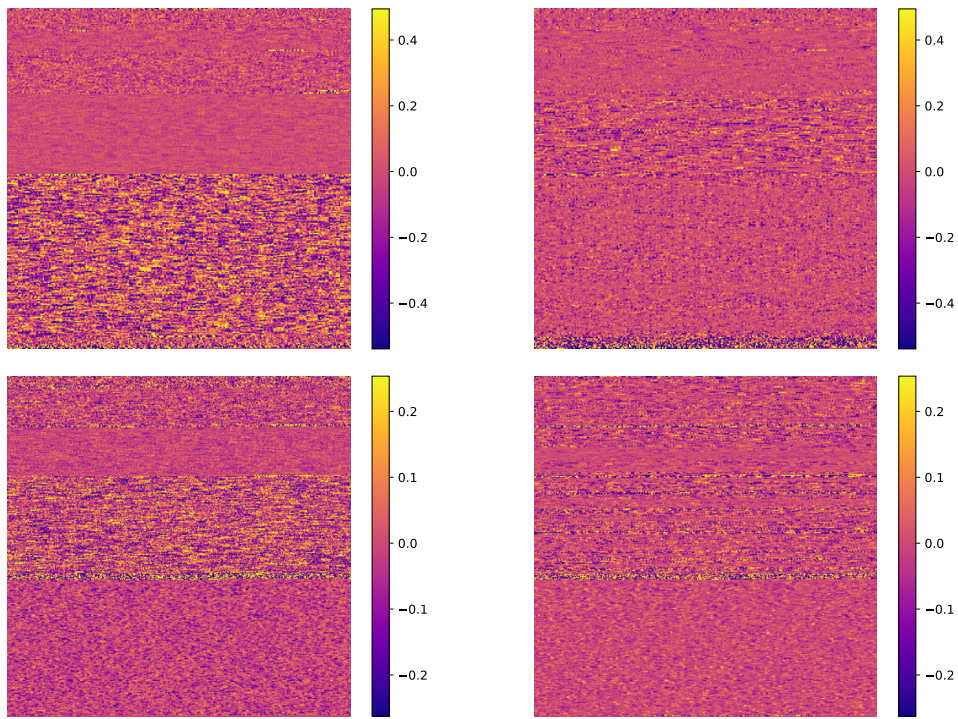

Fig. 5: Visualization of the PRep of a ResNet8 instance in the Single-Architecture setting (Image classification): top, CIFAR-10, bottom, Tiny-ImageNet. The target PRep on the left is given in input to NetSpace, that produces the predicted PRep on the right.

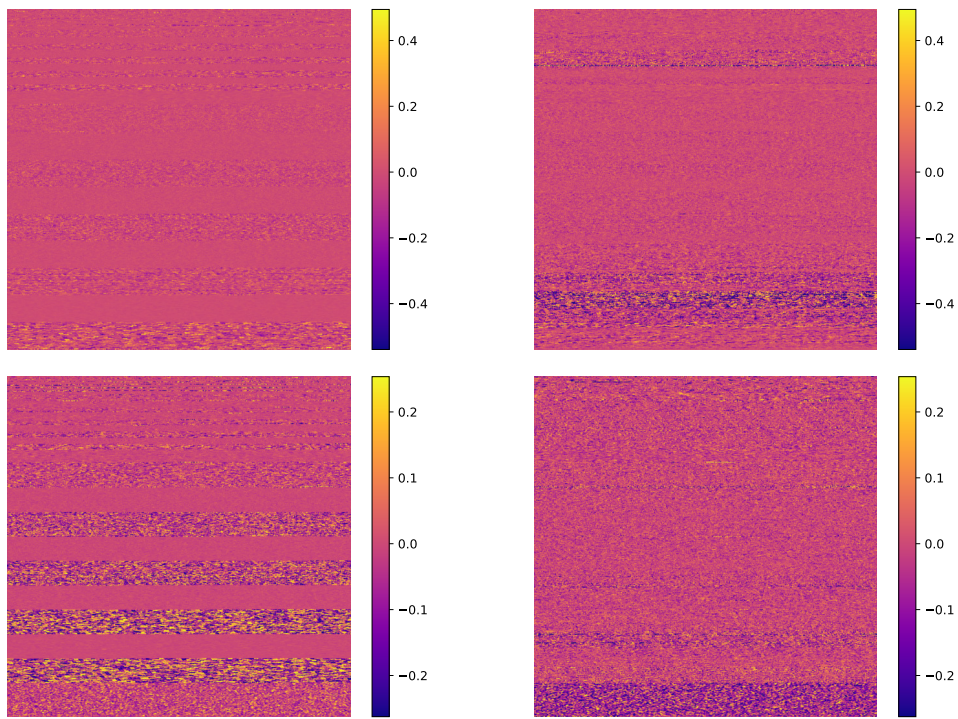

Fig. 6: Visualization of the PRep of a ResNet32 instance in the Multi-Architecture setting: top, CIFAR-10, bottom, Tiny-ImageNet. The target PRep on the left is given in input to NetSpace, that produces the predicted PRep on the right.

the normalized values $\widehat{f}_{i,j} \in \mathbb{R}^C$ at cell $(i,j)$ of its output feature map can be computed as

$$\begin{aligned}\widehat{f}_{i,j} &= W_{BN}\,\tilde{f}_{i,j} + b_{BN} \\ &= W_{BN}\left(W_{conv}\,f_{i,j} + b_{conv}\right) + b_{BN}.\end{aligned} \quad (5)$$

Hence, it is possible to replace every convolutional layer (with weights $W_{conv}$ and $b_{conv}$) followed by a batch norm layer (whose weights can be shaped in $W_{BN}$ and $b_{BN}$ as described above) with a single convolutional layer whose parameters $W$ and $b$ can be computed as:

$$W = W_{BN}\,W_{conv} \quad (6)$$
$$b = W_{BN}\,b_{conv} + b_{BN} \quad (7)$$

## VII. Visualizing Networks as Images

As in our framework network instances are represented as PRep tensors, they can be visualized as images. Thus, in this section we highlight some properties of NetSpace by visualizing input and output instances as images. In particular, following the ResNet8 Single-Architecture (Image classification) and ResNet32 Multi-Architecture trainings, we take some instances from the test set and obtain their PReps along with those of the corresponding instances predicted by NetSpace. As explained in Sec. II of this document, such representations are 2D matrices, that we reshaped to obtain images of form factors amenable to clear visualization. Then, as shown in Fig. 5 and 6, we represent parameters according to a standard colormap.

From these results we can highlight some interesting properties about our framework. Firstly, we observe that PReps predicted by NetSpace are significantly different w.r.t. the input ones: as we did not use any reconstruction loss in the learning objective, NetSpace learnt to predict instances which behave like the input ones but that are different in terms of parameter values. Secondly, we can see that PReps corresponding to different instances can indeed be visualized as different images, which suggests that, perhaps, in future work these images may be used as proxies for neural networks instances, so that training or fine-tuning or distillation may be realized by learning to generate images.


}

\end{document}